\def\year{2018}\relax
\documentclass[letterpaper]{article}
\usepackage{aaai18}
\usepackage{times}
\usepackage{helvet}
\usepackage{courier}
\usepackage{url}
\usepackage{graphicx}
\usepackage{amsmath}
\usepackage{amsfonts}
\usepackage{subfigure}
\usepackage{multirow}
\usepackage{comment}
\usepackage{diagbox}
\usepackage{color}
\nocopyright
\pdfoutput=1

\begin{document}
%
\title{Energy-efficient Amortized Inference with Cascaded Deep Classifiers}
\author{
Jiaqi Guan\\
Tsinghua University \\
UIUC \\
guanjq14@mails.tsinghua.edu.cn \And 
Yang Liu \\
UIUC \\
liu301@illinois.edu \And
Qiang Liu \\
Dartmouth College \\
UT Austin \\
qiang.liu@dartmouth.edu \And
Jian Peng \\
UIUC \\
jianpeng@illinois.edu
}
\maketitle
\begin{abstract}
Deep neural networks have been remarkable successful in various AI tasks but often cast high computation and energy cost for energy-constrained applications such as mobile sensing.  
We address this problem by proposing a novel framework 
that optimizes the prediction accuracy and energy cost simultaneously, thus enabling effective cost-accuracy trade-off at test time. 
In our framework, each data instance is pushed into a cascade of deep neural networks with increasing sizes, 
and a selection module is used to sequentially determine when a sufficiently accurate classifier can be used for this data instance. 
The cascade of neural networks and the selection module are jointly trained in an end-to-end fashion by the REINFORCE algorithm to optimize a trade-off between the computational cost and the predictive accuracy.  
Our method is able to simultaneously improve the accuracy and efficiency by 
learning to assign easy instances to fast yet sufficiently accurate classifiers to 
save computation and energy cost, 
while assigning harder instances to deeper and more powerful classifiers to ensure satisfiable accuracy. 
With extensive experiments on several image classification datasets using cascaded ResNet classifiers, we demonstrate that our method outperforms the standard well-trained ResNets in accuracy but only requires less than 20\% and 50\% FLOPs cost on the CIFAR-10/100 datasets and 66\% on the ImageNet dataset, respectively.

\begin{quote}

\end{quote}
\end{abstract}

\section{Introduction}

The recent advances of deep learning techniques in computer vision, speech recognition and natural language processing have tremendously improved the performance on challenging AI tasks, including image classification \cite{krizhevsky2012imagenet}, speech-based translation and language modeling. Since the first success of deep convolutional neural network in the ImageNet challenge, more complex architectures \cite{simonyan2014very,he2016deep,szegedy2016rethinking,szegedy2017inception} have been proposed to further improve performance, but often at the cost of more expensive computation. However, in many real-world scenarios, such as vision-based robotics and mobile vision applications, we encounter a significant constraint of energy or computational cost for real-time inference. For example, 
mobile applications cast a high demand on fast, energy-efficient inference; 
it is desired to ensure that the majority (e.g.,  90\%) of the users do not feel the latency of the computation, given  that most images are easy to analyze.
This requires new learning methods that are \emph{both accurate and fast}. 

In this paper, we focus on test-time energy-efficient inference of image classification. 
Traditional approaches are usually based on directly 
scarifying accuracy for speed, e.g., by reducing  or compressing well-trained complex neural networks at a cost of loss of accuracy. 
A key observation, however, is that accuracy and cost can be simultaneously improved, and do not necessarily need to scarify for each other; this is because although deeper or more complex networks usually come with higher overall accuracy, 
a large portion of images can still be correctly classified using smaller or simpler networks, 
and the larger networks are necessarily  only for the remaining difficult images.  
Thus, the approach of our work is to jointly train an ensemble of neural networks with different complexity, together with a selection module that adaptively assigns each image to  the smallest neural network that is sufficient to generate high-quality label. 
Unlike traditional learning approaches that learns with constant computation cost, our method learns to \emph{predict both accurately and fast}. 
By the training and using the policy module, our framework yields an efficient amortization strategy, which 
greatly reduce the computational or energy cost in the testing phase with even boosted predictive performance. 

Technically, we frame the training of the neural classifiers and the selection module
into a joint optimization of the training accuracy 
with a constraint on the expected computational cost (in terms of FLOPs cost). 
%
We design the policy module to be a optimal stopping process, 
which sequentially exam the the cascade of neural classifiers with increasing sizes (and hence predictive accuracies), 
and stop at the classifier that optimally trade-off the accuracy and complexity for each given image.
%
Our joint training is performed in an end-to-end fashion by the REINFORCE algorithm \cite{williams1992simple} to optimize a trade-off between the computational cost (in terms of FLOPs cost) and the predictive accuracy as reward signal. 
We perform experiments on the CIFAR and ImageNet classification datasets using a cascade of ResNet classifiers with varying sizes. 
As expected, on the CIFAR datasets, most images are assigned to the smaller networks which are already sufficiently predictive for them, while the remaining difficult images are assigned to larger and more powerful networks. And nearly half of the images are assigned to smaller networks on the ImageNet dataset. Our proposed model outperforms a well-trained accurate deep ResNet classifier in terms of accuracy but only requires less than 20\% and 50\% FLOPs cost on the CIFAR-10/100 and 66\% on the ImageNet dataset, respectively.

\section{Related Work}
There have been a number of existing methods on improving energy efficiency of deep neural networks. Most such techniques focus on simplifying network structure and/or improving basic convolution operations numerically. MobileNet \cite{howard2017mobilenets} uses depthwise separable convolutions to build light weight neural networks. ShuffleNet \cite{zhang2017shufflenet} uses pointwise group convolutions and channel shuffle operation to build an efficient architecture. Other techniques include pruning of connections \cite{han2015deep} and bottleneck structure \cite{iandola2016squeezenet}. In addition to static techniques, Dynamic Capacity Network \cite{almahairi2016dynamic} adaptively assigns its capacity across different portions of the input data by using a low and a high capacity sub-network. Spatially Adaptive Computation Time Networks \cite{figurnov2016spatially} dynamically adjust the number of executed layers for the regions of the image. Anytime Neural Networks \cite{hu2017anytime} can generate anytime predictions by minimizing a carefully constructed weighted sum of losses. Others \cite{li2015convolutional,yang2016exploit} consider cascaded classifiers in object detection to quickly reject proposals that are easy to judge. Conditional computation \cite{bengio2015conditional} and adaptive computation \cite{jernite2016variable,graves2016adaptive} propose to adjust the amount of computational cost by using a policy to select data. Many of these static and dynamic techniques are used in standard deep architectures such as ResNet \cite{he2016deep} and Inception \cite{szegedy2017inception}, usually with a loss of accuracy. Different from these static and dynamic techniques, our method explicitly formulates the test-time efficiency as an amortized constrained sequential decision problem such that the expected computational cost, in terms of FLOPs cost, can be greatly reduced with even improved accuracy by adaptively assigning training examples with various difficulty to their best classifiers. 

\section{Method}
In this section, we first formulate the energy-efficient inference problem as an optimization with amortized constraint. Then we reduce it to a sequential decision process and proposed a solution based on REINFORCE algorithm. Finally, we introduce the details of implementation like classifier structure, policy module structure used in the experiments.

\begin{figure*}[h]
\centering
\includegraphics[width=1.0\textwidth]{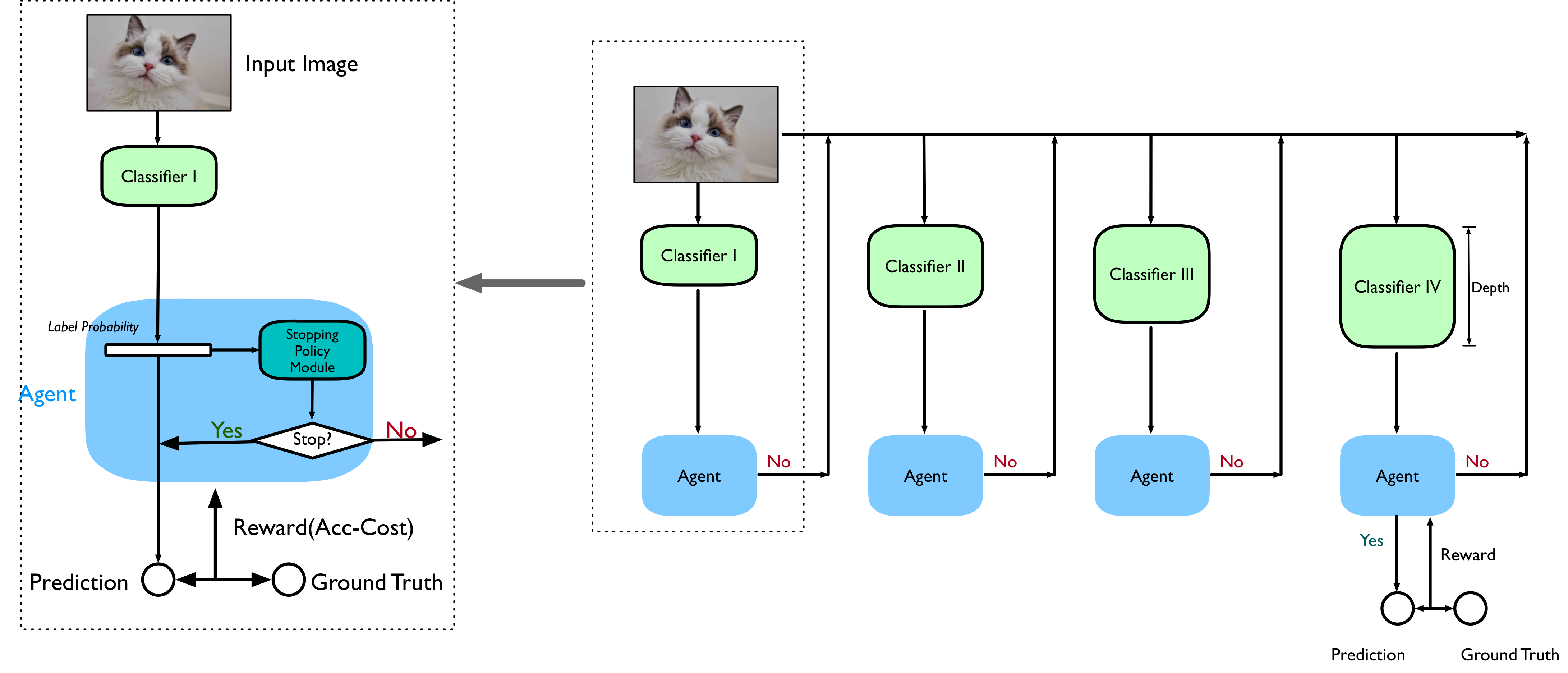}
\caption {\textbf{Our proposed model:} Given an image in the dataset, starting from smallest model, our agent will decide whether to move to the next deeper model. If we decide to stop at a classifier, we predict the label based on the classifier. Finally, the agent will receive a reward as we described in the method section. Inside our agent, a stopping policy module takes label probability of a classifier's top layer as input and decides whether to stop or continue.}
\label{fig:model}
\end{figure*}

\subsection{Energy-constrained Inference of Cascaded Classifiers}
Classifiers, such as neural networks, are often more accurate with deeper or more complex architectures. However, the high computational or energy cost of complex networks are prohibitive for fast, real-time inference in applications deployed on mobile devices. If we have a cascade of classifiers with different sizes, it is possible to select the smallest, yet sufficiently powerful classifier for each input data to achieve both efficiency and accuracy simultaneously. This introduces our main problem: 
\emph{Given a cascade of neural classifiers with different accuracies and cost,
how to train them jointly together with an efficient selection mechanism 
to assign each data instance to the classifier that optimally trade off accuracy and cost? }

Specifically, suppose we have $K$ classifiers $\{C_k\}_{k=1}^K$ with different energy cost $\{\mathcal{F}_k\}_{k=1}^K$. The energy cost $\mathcal F_k$ is assumed to correlate with the predictive capacity of classifiers, and can be, for example, a normalized value of FLOPs or the number of layers in neural network classifiers. Given an input $x$, we denote by $y$ its true label and 
$\hat y \sim C_k(\cdot | x)$  the label predicted by classifier $C_k$. 
In addition, we denote by $\Pi(k|x)$ a randomized policy that decides the probability of assigning input $x$ to classifier $C_k$. 
Our target is to jointly train all classifiers $\{C_k\}$ and the policy $\Pi(k|x)$ to minimize the expected loss function under the constraint that the expected energy cost should be no larger than a desired budget $\mathcal{B}$, that is, 
%
\begin{align*}
    &\max_{\Pi, \{C_t\}_{t=1}^K}{\mathbb{E}_{(x, y) \sim \mathcal{D}, k_x\sim \Pi(\cdot|x), \hat y \sim C_{k_x}(\cdot|x)}\left[-\mathcal{L}\left(\hat y ,y\right)\right]} \\
    &s.t ~~~ \mathbb{E}_{(x,y) \sim \mathcal{D}, k_x\sim \Pi(\cdot|x)}\left[\mathcal{F}_{k_x}\right] < \mathcal{B},
\end{align*}
where $k_x$ denotes the (random) classifier ID assigned to $x$. 
Further, we can reform the constrained optimization 
into an unconstrained optimization of a penalized cost function: 
\begin{align*}
    &\max_{\Pi, \{C_t\}_{t=1}^K} {\mathbb{E}_{(x, y) \sim \mathcal{D}, k_x\sim \Pi(\cdot|x), y^\prime \sim C_{k_x}(\cdot|x)}\left[-\mathcal{L}\left(y^\prime ,y\right) -\alpha \mathcal{F}_{k_x}\right]}
\end{align*}
where $\alpha$ controls the trade-off between the predictive loss function and the energy cost. 
There is an (implicit) one-to-one map between the budget constraint $\mathcal B$ and the penalty coefficient $\alpha$ under which these two forms are equivalent in duality. We will use the penalized form in our experiments for its simplicity. 

\subsection{Energy Efficient Inference via Optimal Stopping}
The design of the selection module $\Pi$ plays an critical role in our framework. 
It should (i) get access to and efficiently leverage
the information of the classifiers $\{C_k\}$ to make reasonable decisions, 
and (ii) be computationally efficient, e.g., at least avoiding  brute-forcely eliminating all the $K$ classifiers and selects one the with largest confidence.  
We propose to resolve this challenge by framing $\Pi$ into a $K$-step optimal stopping process. At each time step $t$, we introduce a stopping policy module, which takes some feature $s_t(x)$ related to classifier $C_t$, and output a stopping probability $\pi_t(s_t(x))$  with which we decide to stop at the $t$-th classifier and take it as the final predictor for input $x$. Otherwise, we will move to a deeper classifier and repeat the same process until it reaches the deepest one. In this way, the overall probability of selecting at the $k$-th classifier is 
$$
\Pi(k | x) = \pi_k(s_k(x)) \prod_{t=1}^{k-1} (1 -\pi_t(s_t(x))). 
$$
Suppose we finally stop at the $k$-th classifier,  our agent receives a reward consisting of two parts: the  loss function for prediction using the selected classifier , i.e., $\mathcal L(\hat y, y)$ where $\hat y \sim C_k(\cdot | x)$,  and the energy cost accumulated from the first classifier till current one, i.e.,  $\sum_{t=1}^k \mathcal{F}_t$. In practice, we also incorporate the accumulated computational cost of the stopping policy $\pi_t$ in each $\mathcal{F}_t$. 
Importantly, once we stop at the $k$-th classifier, we no longer run the classifiers that are more expensive than $k$, which significantly saves the computational cost. 
Overall, this defines the following the reward signal: 
\begin{align}\label{equ:px}
R(k,x,y,\hat y) = -\mathcal{L}\left(\hat y , y\right) - \alpha \sum_{t=1}^{k-1}{\mathcal{F}_t}
\end{align}
To recap, our decision module is framed as a Markov decision process consisting of the following components: 
\begin{itemize}
    \item \textbf{Observation:} The stopping probability $\pi_{t}(s_t(x))$ at the $t$-th step depends on a feature $s_t(x)$ which should represent the confidence level of the $t$-th classifier  $C_t$.  In this work, we simply use the output probability as the observation at each step, that is, $s_t(x) = C_t(\cdot | x)$. 
    \item \textbf{Action:} Based on the output probability of the current classifier, our stopping policy module decides to stop at the current step with probability $\pi_t(s_t(x))$. If it finally stops at the $k$-th step, we use the current model $C_k$ to predict the label, that is, $\hat y \sim {C}_k(\cdot| x)$. 
    \item \textbf{Reward:} After finally stopping at one classifier, the agent receives a reward signal shown in Eq \eqref{equ:px} consisting of both the negative loss function for prediction and the accumulated energy cost from the first step. In this paper, we use a normalized FLOPs count as the cost. 
\end{itemize}
Assume the stopping probabilities $\{\pi_t\}$ and classifiers $\{C_t\}$ are parameterized by 
$\theta = \{\theta^{\pi_t}, \theta^{C_t}\}_{t=1}^K$. 
Our final goal is to find the optimal $\theta$ is to maximize the expected return, by unrolling the conditional distributions defined by the entire policy:
\begin{align*}
J(\theta) &= \mathbb{E}_{(x,y) \sim \mathcal{D}}\big[\mathbb{E}_{k \sim \Pi(\cdot | x), \hat y \sim C_k(\cdot|x)}{R(k, x, y, \hat y)} \big] \\
&=\mathbb{E}_{(x,y) \sim \mathcal{D}}\big[\sum_{k=1}^{K}{\prod_{t=1}^{k-1}{(1 - \pi_t(s_t(x);{\theta}))}} \\
&\cdot \pi_k(s_k(x); \theta) \cdot \sum_{\hat y} C_k(\hat y|x;{\theta}) \cdot R(k, x, y, \hat y) \big].
\end{align*}

\subsection{Solving by REINFORCE}
To solve this optimal  stopping problem, we apply the well-known REINFORCE algorithm \cite{williams1992simple} by rolling out each individual sample $(x,y)$ according to the current parameter and derive the policy gradient in following form: 
\begin{align*}
\widehat{\nabla_{\theta}{J}} = 
 \nabla_\theta\big [  \sum_{t=1}^{k-1}{{\log(1 - \pi_t(s_t(x); {\theta}))}}+{\log(\pi_k(s_k(x);{\theta}))}\\ + {\log(C_k(\hat{y}|x;{ \theta}))} \big ]   
\cdot  R(k,x,y,\hat{y}).
\end{align*}
Moving further, we introduce a \textit{baseline} $b$ to reduce the variance in the estimated policy gradient, resulting the following gradient estimation:



\begin{align*}
\widehat{\nabla_{\theta}{J}} = 
 \nabla_\theta\big( \sum_{t=1}^{k-1}{{\log(1 - \pi_t(s_t(x); {\theta}))}}+{\log(\pi_k(s_k(x);{\theta}))}\\ + {\log(C_k(\hat{y}|x;{ \theta}))} \big)  
\cdot (R(k,x,y,\hat{y})-b) 
\end{align*}
 \noindent where the \emph{baseline} $b$  is chosen by minimizing the variance of the gradient estimator on a mini-batch of the training data. 

\subsection{Cascaded classifiers using ResNet}
In this paper, we use image classification for benchmarking our method. Deep residual network \cite{he2016deep} has been widely used in image classification field since it was proposed. 
The ResNet architecture we use are specified as follows: 
The first two layers of ResNet are a convolution layer and a pooling layer with a total stride of 4, while for small images, such as images in CIFAR-10 and CIFAR-100 dataset, it can be only a convolution layer with stride 1. 
Then, a sequence of blocks is stacked together. 
Each block has different numbers of units and each residual unit performs the residual learning, which has a form $y = x + F(x, {W_i})$, where x is called shortcut connection and F(x) is called residual function. The residual function we use is basic residual, which is usually used in the scene of small input and not very deep neural networks. It consists of two $3\times3$ convolution layers that both have equal input and output channels. Finally, the output of last unit will be passed through a global average pooling layer \cite{lin2013network} and a fully-connected layer to obtain the logits of the class probabilities.

We choose ResNet as our model's baseline because we can easily build a sequence of networks from shallow to deep by adjusting the number of units in each block. Generally, the deeper network has the better prediction performance, though having more computational cost. Then, we can attach the policy network to this sequence of networks to achieve our algorithm, which will be described in detail in the following sub-section.

Our ResNet is implemented in pre-activation \cite{he2016identity} version, in which each convolution layer is preceded by a batch normalization layer \cite{ioffe2015batch} and a ReLU non-linear unit. In addition, after each block, the feature map size is halved and the number of filters is doubled, which follows the Very Deep Networks design \cite{simonyan2014very} and ensures all units have equal computational cost. 

\section{Implementation and Experimental Setting}

\noindent\textbf{Datasets.} As a proof of concept, we implement a cascade of deep neural network classifiers on three image classification datasets, including CIFAR-10, CIFAR-100 \cite{krizhevsky2009learning}, and ImageNet32x32 \cite{chrabaszcz2017downsampled}.  These three datasets consist of 32x32 RGB colored images. The CIFAR-10 and CIFAR-100 datasets both have 50000 training images and 10000 test images, with 10 classes and 100 classes respectively. The ImageNet32x32 dataset is a down-sampled variant of origin ImageNet dataset \cite{deng2009imagenet}, which contains the same classes (1000 classes) and the same number of images (1.2 million training images and 50000 test images) with a reduced resolution of 32x32 pixels.

\begin{table}[h]
\centering
\caption{{\bf The ResNet structure  used and their FLOPs.} For the CIFAR datasets, our classifier cascade consists of ResNets with different layers; For the ImageNet dataset, it is ResNet with 40 layers of different widths. }
\scalebox{.95}{
\begin{tabular}{|c|c|c|c|c|}
\hline
\multicolumn{3}{|c|}{CIFAR} & \multicolumn{2}{c|}{ImageNet32x32} \\
\hline
Layer & 
\!\!\!\!\! \!\!\!\!\begin{tabular}{c}Unit\\ Number \end{tabular} \!\!\!\!\!\!\!\!\!
& FLOPs(M)
& \!\!\!\!\! \!\!\! \begin{tabular}{c} Width \\ Multiplier \end{tabular} \!\!\!\!\!\!\!\! 
& FLOPs(M)\\
\hline
8 & [1, 1, 1] & 14.86 & 1 & 85.64\\
\hline
20 & [3, 3, 3] & 43.17 & 1.5 & 192.36\\
\hline
32 & [5, 5, 5] & 71.48 & 2 & 341.68	\\
\hline
56 & [9, 9, 9] & 128.11 & 3 & 768.13\\
\hline
110 & [18, 18, 18] & 255.51 & 4 & 1364.97\\
\hline
\end{tabular}
}
\label{tab:base_resnet_flops}
\end{table}

\begin{table}[h]
\centering
\caption{{\bf Comparing with Static ResNet classifiers.} We compare our model to well-trained, static ResNet classifiers with different sizes. 
In our method, the hyperparameter $\alpha$ is selected to match the accuracy of our method with that of the best static ResNet. 
Our model achieves not only lower error but also significantly less cost on all three datasets.}
\begin{tabular}{c|c|c}
\hline
\multicolumn{3}{c}{\textbf{CIFAR-10}}\\
\hline
Model & Error & Relative FLOPs\\
\hline
ResNet-8 	& 12.33\% & 5.82\%	\\
ResNet-20 	& 9.00\% & 16.90\%	 	\\
ResNet-32 	& 8.40\% & 27.98\%		\\
ResNet-56 	& 7.70\% & 50.14\%	\\
ResNet-110 	& 7.38\% & 100.00\%	\\
\hline
Ours		& \textbf{7.20\%} & \textbf{19.20\%	} 	\\
\hline
\hline
\multicolumn{3}{c}{\textbf{CIFAR-100}}\\
\hline
Model & Error & Relative FLOPs\\
\hline
ResNet-8 	& 39.98\% 	& 5.82\%	 		\\
ResNet-20 	& 33.13\% 	& 16.90\%			\\
ResNet-32 	& 31.56\% 	& 27.98\%			\\
ResNet-56 	& 30.38\% 	& 50.14\%			\\
ResNet-110 	& 28.63\% 	& 100.00\%			\\
\hline
Ours		& \textbf{27.86\%}&  \textbf{49.33\%	}	\\
\hline
\hline
\multicolumn{3}{c}{\textbf{ImageNet32x32 (Top-5 Error)}} \\ 
\hline
Model & Error & Relative FLOPs\\
\hline
ResNet40-1 	&	39.72\%		&	 6.27\% \\
ResNet40-1.5		&	32.76\%		&	14.09\% \\
ResNet40-2 	&	29.64\%		&	25.03\% \\
ResNet40-3 	&	24.67\%		&	56.27\% \\
ResNet40-4 	&	22.22\%		& 	100.00\% \\
\hline
Ours		& \textbf{22.21\%}		& 	\textbf{66.22\%}	\\
\hline
\end{tabular}
\label{tab:results}
\end{table}


\noindent\textbf{Neural classifier specification.} To construct a cascade of neural network classifiers, we take the standard design of the ResNet architecture \cite{he2016deep} to build a sequence of ResNets with nearly exponentially increasing depths. In this cascade, each ResNet classifier starts with a $3\times3$ convolution layer with 16 filters, followed by three blocks of residual units. Each unit consists of two $3\times3$ convolution layers. In the second and third blocks, the number of filters is doubled and the size of feature map is halved at the first unit. The numbers of units in each block are set to 1, 3, 5, 9, 18 to build this sequence of ResNets, with 8, 20, 32, 56, 110 layers respectively. For the ImageNet32x32 dataset, we adopt a width multiplier $k$ following Wide-ResNet \cite{zagoruyko2016wide} to  increase the capacity of individual classifiers by changing the number of filters. We set the width multipliers to 1, 1.5, 2, 3, 4 and the number of convolution layers to 40, so that the capacity or FLOPs of the ResNet classifiers are approximately exponentially increasing. We notice that this adoption works better than the original ResNet setting, due to the larger volume and higher diversity of the ImageNet dataset. 

Here our design ensures that the depth (and hence the computational complexity) of the cascade of neural network increases exponentially. This ensures that we do not waste significant computation resource in examining the smaller networks. To be more specific, 
assume the computational cost of the network classifiers are $b^{k},$ $k =1,\ldots, K$, where $b>1$, 
then the cost when we stop at the $k$-th classifier is $\sum_{\ell=1}^k b^l  \leq \frac{b}{b-1} b^k$, which is at most $\frac{b}{b-1}$ times of $b^k$, the cost incurred when we select the $k$-th classifier by oracle, without examining any of the weaker classifiers. 

\noindent\textbf{Policy module specification.} 
 The stopping policy module is constructed using  
 three fully-connected layers, with 64 hidden neurons each for both CIFAR-10 and CIFAR-100, and with 256 hidden neurons for the ImageNet32x32 dataset. Each fully-connected layer is followed by a ReLU non-linear unit except the last layer. Finally, the fully-connected layers are followed by a softmax function which outputs the stopping probability.  The input of this module is the label probability output from individual ResNets. This stopping policy module has nearly negligible computational cost ($\sim$0.01M FLOPs on the CIFAR datasets and $\sim$0.1M FLOPs on the ImageNet dataset), compared to that of the ResNets (see Table \ref{tab:base_resnet_flops}). We notice that other features, such as top-layer convolutional filters, would greatly increase the computational cost of this stopping policy module and have a lower classification accuracy. 

\noindent\textbf{Other implementation details.} During the training phase, we adopt the standard data augmentation procedure \cite{lee2015deeply,he2016deep} on all three datasets: padding each side of the images by four zeros and randomly cropping a 32x32 image; randomly flipping left to right. For all the experiments, we use  stochastic gradient descent with a momentum of 0.9 for the policy optimization. The learning rate schedule and the mini-batch size are set to be the same as in the original ResNet \cite{he2016deep} for the gradients associated. The learning rate for the stopping policy module is set to be  0.05 and an exponential decay with a factor of 0.9 is applied every two epochs according to internal cross-validation within the training data.

To evaluate the classification performance, we use the top-1 accuracy for the CIFAR-10 and CIFAR-100 datasets and an additional top-5 accuracy for the ImageNet32x32 dataset. For training, we also use the top-1 accuracy in the reward signal for the CIFAR-10 and CIFAR-100 datasets while the top-5 accuracy for the ImageNet32x32 dataset. We use the per-image number of floating-point operations (FLOPs) to measure the computational cost, with multiplications, additions  and multiply-add operations considered.  
The network's size and FLOPs of individual ResNets are shown in Table \ref{tab:base_resnet_flops}. 





\begin{figure}[h]
\centering
\includegraphics[width=0.4\textwidth]{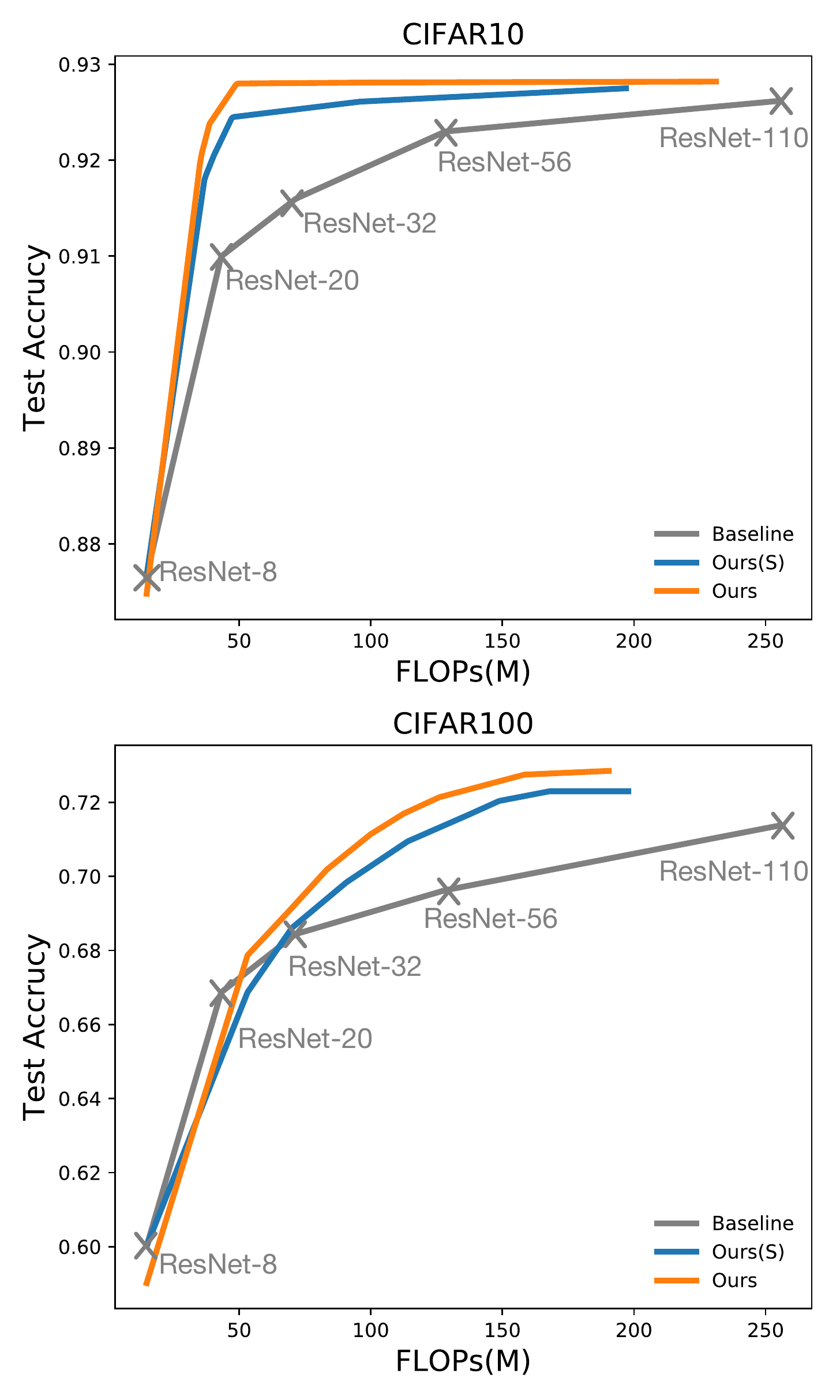}
\caption{{\bf Results on CIFAR-10/100}: The x-axis denotes the millions of FLOPs and y-axis denotes  the corresponding accuracy obtained by the static ResNet (gray), our model  (orange), and the simplified version of our model (blue), respectively. It is obvious that our models outperform the static ResNet under the same energy cost on a large spectrum.}
\label{fig:analysis_cifar}
\end{figure}

\begin{figure*}
\centering
\includegraphics[width=0.85\textwidth]{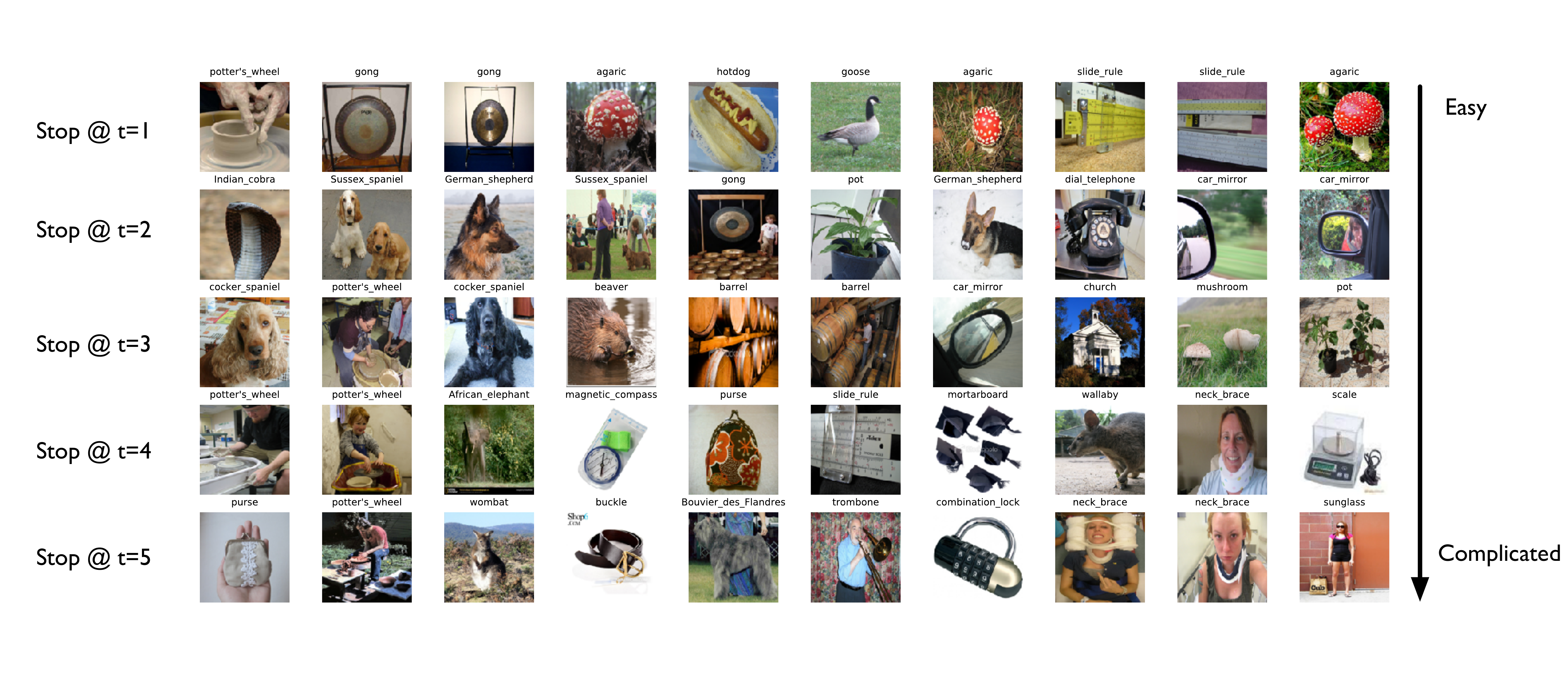}
\caption{{\bf Visualizing ImageNet}: Row $i$ contains the top-10 images stopped at the $i$-th classifier, sorted by the corresponding selection probabilities. We can visually see that the images become increasingly challenging to classify from top to bottom.}
\label{fig:imagenet_vis1}
\end{figure*}

\begin{figure*}
\centering
\includegraphics[width=0.75\textwidth]{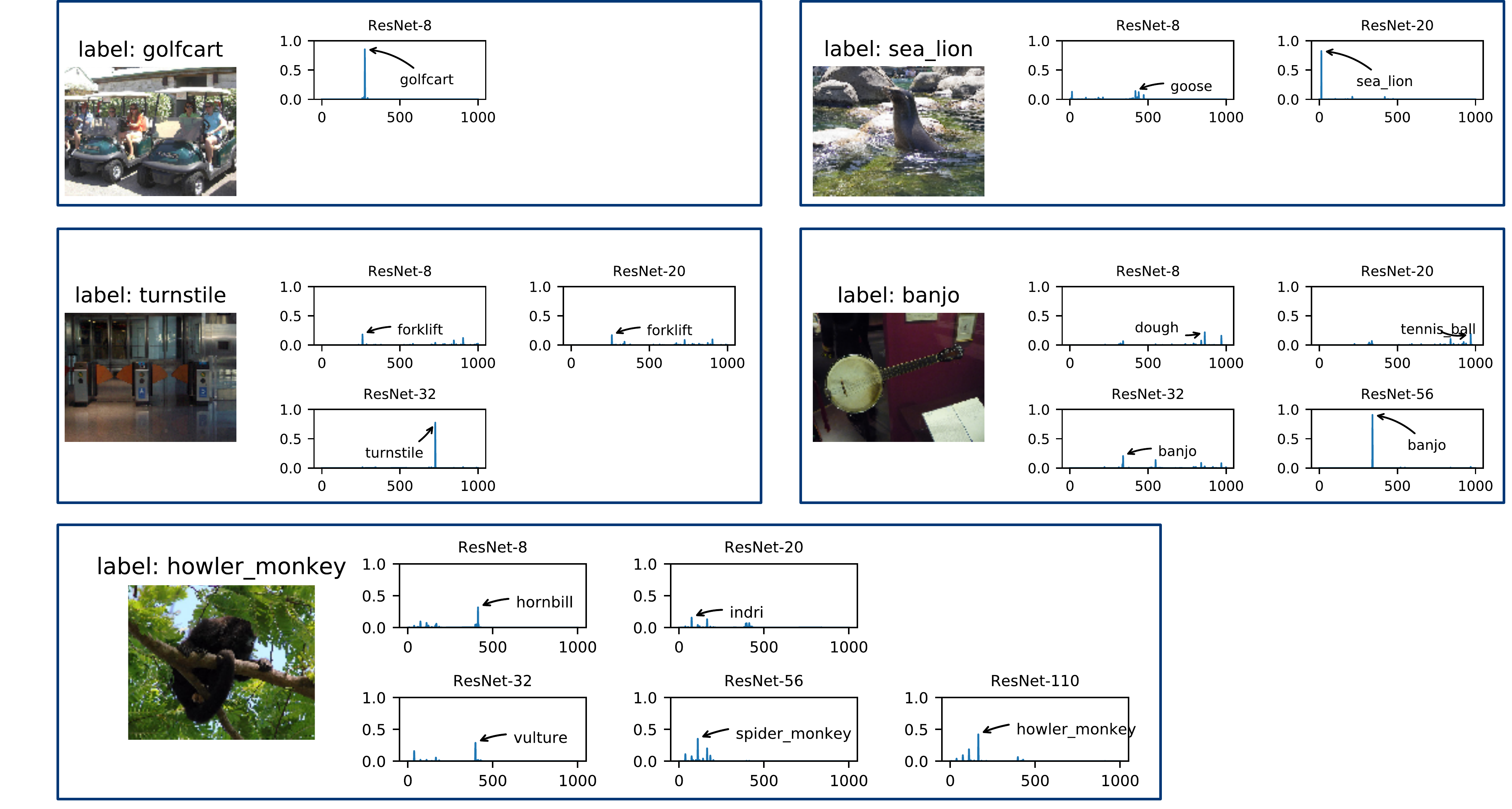}
\caption{{\bf Example images and their predicted label probabilities}: Five images from ImageNet assigned to different ResNet classifiers, with their label probability distributions given by the classifiers they visited.} 
\label{fig:imagenet_vis2}
\end{figure*}

\begin{figure*}
\centering
\includegraphics[width=0.85\textwidth]{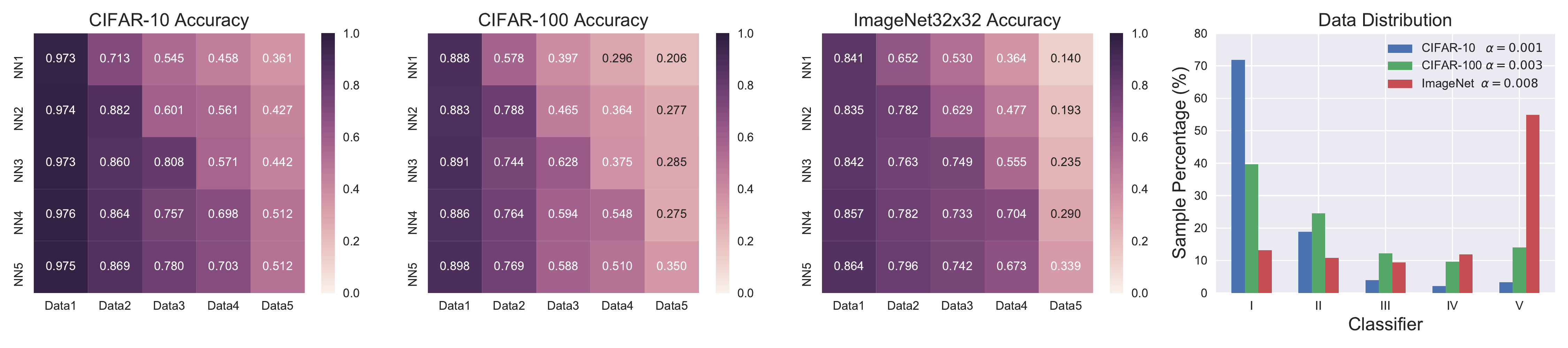}
\caption{{\bf Accuracy distribution of classifiers in the cascade}: The first three figures plot accuracy distributions on the CIFAR-10, CIFAR-100 and ImageNet32x32 datasets. The $i$-th row and $j$-th column denotes the average accuracy predicted by the $i$-th classifier of samples assigned to the $j$-th classifier. The fourth figure is the distribution of test images on individual classifiers, where x-axis indexes five classifiers and y-axis denotes the proportion of samples eventually assigned to the corresponding classifier.}
\label{fig:analysis}
\end{figure*}

\section{Results}
We compare our model to well-trained ResNet classifiers with 8, 20, 32, 56 and 110 layers,  respectively, whose architectures are constructed in the same way as the classifiers in our cascaded model.  We vary the hyperparameter $\alpha$, the coefficient of the energy cost in the reward signal,  in the range of $10^{-4}$ to $10^{-2}$ to demonstrate the trade-off between the computational cost (FLOPs) and accuracy.  The comparisons on CIFAR-10 and CIFAR-100 are shown in Figure \ref{fig:analysis_cifar} and Table \ref{tab:results}. 
In Figure ~\ref{fig:analysis_cifar}, the gray curve shows the performance of the static ResNet classifiers with difference numbers of layers, and the orange curve shows the performance of our model with different values of $\alpha$ . Clearly, our model achieves not only better classification accuracy but also higher cost effectiveness. Our model can achieve 0.18\% higher accuracy with only 19.20\% FLOPs on the CIFAR-10 dataset, compared to the best performing static ResNet110 classifier. On the CIFAR-100 set, our model obtains 0.77\% higher accuracy with only 49.33\% FLOPs, compared to the best static ResNet110 classifier. It is worth noting that the efficiency improvement is more prominent on the CIFAR-10 than the CIFAR-100, due to the fact that CIFAR-100 is a more challenging dataset so that deeper classifiers are more frequently required to distinguish similar labels. 
Similarly, the result on the ImageNet32x32 dataset (Table \ref{tab:results}) shows that our model can achieve almost the same top-5 accuracy compared with the largest static ResNet classifier but only requires 66.22\% computational cost.

As another baseline, we have implemented a simplified version of our model (denote as ``Ours (S)'' in Figure~\ref{fig:analysis_cifar}), in which we only train the stopping policy model to sequentially decide which the classifiers to use, and these classifiers within the cascade are pre-trained and fixed. The blue curve in Figure \ref{fig:analysis_cifar} indicates that this version can also outperform the best ResNet110 classifier as it dynamically decides the smallest classifier that is sufficient for a input image. Our jointly trained model shows further improved performance, especially when the amortized FLOPs required is small, as our model also determines which images to use for classifiers in the cascade during training, compared to this simplified version.


%


\subsection{Further analysis}
We further investigate how our model works exactly by visualizing representative samples assigned to different classifiers. Figure \ref{fig:imagenet_vis1} shows the top 10 ImageNet32x32 images assigned to the five classifiers sorted by the selection probability to each classifier. 
From this figure, 
we can see that many of the images correctly classified by the shallowest classifier 
 indeed looks easy, such as \emph{gong} and \emph{agaric}, 
 while the images that require more powerful classifiers looks more challenging visually, such as \emph{potter's wheel}. 

Figure \ref{fig:imagenet_vis2} shows five images stopped at various classifiers and their label probability distributions given by the classifiers they visited before stopping. 
Taking the third samples as examples, we can see that the \emph{turnstile} is first confused with \emph{forklift} by the first two classifiers, but is correctly classified by the third classifier on which it stops.   
In the  fifth image, \emph{howler monkey} is identified by the fifth classifier, but is 
wrongly predicted by all the first four classifiers as other classes, including \emph{spider monkey} which is indeed easily confused with the true label. 





We can divide the testing datasets into 5 subsets according to which classifier an image is assigned to by the selection module. This means that subset $i$ contains all the images assigned to the $i$-th classifier in the cascade. 
Figure \ref{fig:analysis} shows the average accuracy of the different subsets on different networks.  
We can see that for the data assigned to classifier $i$, the accuracy at classifier $i$ is consistently higher than the accuracies of classifier $1$ to $i-1$ on all three datasets (i.e., the diagonals are larger than the upper triangular elements), suggesting that the selection modules successfully identify more accurate classifiers. 
Interestingly, we find that the accuracy does not always increase when ResNet becomes deeper. 
For example, on the relatively simple datasets, CIFAR-10 and CIFAR-100, the accuracies of subset $i$ at classifier $i$ ($i$=2,3,4,5) are even higher than accuracies at classifier $i+1$ to $5$. 
This is because the REINFORCE algorithm  distributes only harder images to the deeper ResNets, making them less accurate on the easier images. 

The rightmost panel of Figure \ref{fig:analysis} shows the proportions of the 5 subsets in different datasets. 
We can see that in CIFAR-10 (blue) and CIFAR-100 (green), most images are easy to classify and are assigned to the smaller classifiers, while ImageNet is more difficult and  a majority of it is assigned to the largest classifier. 
Even in ImageNet, our method successfully identifies a large portion of easier images, 
and hence obtain better average FLOPs than the biggest static ResNet. 
%
%


\section{Conclusion}
In this work, we propose an energy-efficient model by cascading deep classifiers with a policy module. The policy module is trained by REINFORCE to choose the smallest classifier which is sufficient to make accurate prediction for each input instance. Tested on image classification, our model assigns a large portion of images to the smaller networks and remaining difficult images to the deeper models when necessary. In this way, our model is able to achieve both high accuracy and amortized efficiency during test time. We evaluate our energy-efficient model on the CIFAR-10, CIFAR-100 and ImageNet32x32 datasets. It obtains nearly the same as or higher accuracy than well-trained deep ResNet classifiers but only requires approximately 20\%, 50\% and 66\% FLOPs cost respectively. With a spectrum of computational cost parameter $\alpha$ values, our model achieves different trade-offs between amortized computational cost and predictive accuracy.

\bibliographystyle{aaai}
\bibliography{reference}
\end{document}